\documentclass{article}

\usepackage{graphicx}
\usepackage{floatrow}
\usepackage[nodisplayskipstretch]{setspace}
\usepackage{wrapfig, framed, caption}
\usepackage{pdfpages}
\usepackage{subcaption}
\usepackage{enumerate}
\usepackage{enumitem}
\usepackage{float}
\usepackage{cite}
\usepackage{graphicx}
\usepackage{pgfplots}
\usepackage{tikz}
\usepackage{epstopdf}
\usepackage{amsmath}
\usepackage{tikz}
\usetikzlibrary{positioning}

\usepackage{algorithm}

\usepackage{algpseudocode}

\usepackage[final]{nips_2016}

\usepackage{graphicx}
\usepackage[utf8]{inputenc} 
\usepackage[T1]{fontenc}    
\usepackage{hyperref}       
\usepackage{url}            
\usepackage{booktabs}       
\usepackage{amsfonts}       
\usepackage{nicefrac}       
\usepackage{microtype}      
\usepackage{amsfonts}
\usepackage{natbib}
\title{A Noise-Filtering Approach for Cancer Drug Sensitivity Prediction}

\author{
  Turki ~Turki$^{1,2}$ \\
  $^1$King Abdulaziz University\\
  \texttt{tturki@kau.edu.sa} \\
\parskip 0pt  
  \And
  Zhi Wei$^{2}$ \\
  $^2$New jersey Institute of Technology \\
  \texttt\{ttt2,zhiwei\}@njit.edu \\
}

\begin{document}

\maketitle

\begin{abstract}
\parskip 0pt  
Accurately predicting drug responses to cancer is an important problem hindering oncologists' efforts to find the most effective drugs to treat cancer, which is a core goal in precision medicine. The scientific community has focused on improving this prediction based on genomic, epigenomic, and proteomic datasets measured in human cancer cell lines. Real-world cancer cell lines contain noise, which degrades the performance of machine learning algorithms. This problem is rarely addressed in the existing approaches. In this paper, we present a noise-filtering approach that integrates techniques from numerical linear algebra and information retrieval targeted at filtering out noisy cancer cell lines. By filtering out noisy cancer cell lines, we can train machine learning algorithms on better quality cancer cell lines. We evaluate the performance of our approach and compare it with an existing approach using the $\textit{Area Under the ROC Curve}$ (AUC) on clinical trial data. The experimental results show that our proposed approach is stable and also yields the highest AUC at a statistically significant level.
\end{abstract}

\section{Introduction}
\parskip 0pt        
Cancer has a significant impact on public health worldwide and is the second leading cause of death in the US \cite{siegel2015cancer}. In 2016, the American Cancer Society \cite{amcstarver2016cancer} predicted that 1,685,210 new cancer cases will be diagnosed, resulting in 595,690 deaths attributable to cancer in the US. Many of these cancer patients respond differently to the same cancer drug (i.e., during chemotherapy). These response differences are attributable to environmental (i.e., external) factors such as tobacco, infectious organisms, and an unhealthy diet, as well as to genetic (i.e., internal) factors such as inherited genetic mutations, hormones, or immune conditions, and cancer cell heterogeneity, all of which make cancer drug discovery very difficult \cite{kamb2007cancer,marx2015cancer,bedard2013tumour,roden2002genetic}. Because of the significant numbers of deaths associated with cancer, its study has attracted the attention of researchers from numerous domains including computational biology, machine learning, and data mining \cite{libbrecht2015machine,covell2015data,7591437embc2016}.

\ \ \ \ Many existing drug sensitivity prediction algorithms do not take sample quality into consideration \cite{costello2014community,geeleher2014clinical}, which degrades their performance in the real world. Cell lines of poor quality exist, especially when cell lines are not screened against all of the compounds \cite{yadav2015drug}. These existing approaches fail to remove the poor cell lines, which correspond to noisy samples in machine learning terms, and this failure leads to the degraded performance of machine learning algorithms \cite{mohri2012foundations}.

\ \ \ \ $\textit{Contribution}$: In this paper, we propose a learning approach that removes the noisiest cell lines, allowing a model to be learned from better quality cell lines to improve the predictive performance. Our proposed approach consists of three steps. First, we calculate a distance matrix that corresponds to all the inner products of the rows of a given matrix constructed using the Manhattan distance of the training input. Second, we adopt technique from linear algebra to project the training input on the eigenvectors of the distance matrix to yield transformed training input that corresponds to feature vectors with noise-filtered features. Third, we adopt information retrieval technique \cite{Manning:2008:IIR:1394399} to retrieve (i.e., select) a subset of better quality cell lines with the associated drug responses from the training set using the degrees between the training input cell lines and the corresponding transformed training input cell lines, where smaller degrees denote better quality cell lines. Then, we apply a learning algorithm to the better quality training set to induce (i.e., learn) a model used for prediction on the test set. The learning algorithms considered here are support vector regression and ridge regression \cite{geeleher2014clinical,scholkopf2002learning}. We excluded other learning algorithms such as random forests \cite{breiman2001random} because of their poor performance, as in  \cite{geeleher2014clinical}. As our experimental results show, the proposed approach outperforms the baseline prediction algorithms proposed by Geeleher et al.  \cite{geeleher2014clinical}.

\quad $\textit{Outline}$: Section 2 describes the details of our proposed approach. Section 3 reports the experimental results, including the evaluation of our proposed approach against the baseline prediction algorithms on clinical trial data pertaining to breast cancer and multiple myeloma. Section 4 concludes the paper.
\section{The Proposed Approach}
\lineskip
\parskip 
Figure 1 outlines the proposed approach, which works as follows Consider the gene expression profiles denoted by$\text{X}\in {{\mathbb{R}}^{m\,\times \,n}}$., which consists of $\textit{m}$ cell lines (i.e., samples) and $\textit{n}$ genes. $\text{Y}\,={{({{y}_{1}}\text{,}...,{{y}_{m}})}^{T}}\,$ consists of the corresponding real-value drug responses to $\text{X}$. A training set is define as $\text{S= }\!\!\{\!\!\text{ (}{{\text{x}}_{i}},{{y}_{i}})\}_{i=1}^{m}$, ${{\text{x}}_{i}}\in \,\,\text{X}\,\text{and}\,\,{{y}_{i}}\in \text{Y}\text{.}$ A test set is defined as $\text{T}=\{\text{x}_{i}^{'}\text{ }\!\!\}\!\!\text{ }_{i=1}^{p},$where $\text{x}_{i}^{'}\in {{\mathbb{R}}^{n}}.$ (B) We compute the distance matrix D as
\begin{align}
\text{D}=\text{L}{{\text{L}}^{T}}
\end{align}
where
\begin{align}
\text{L}\,=\,{{[l({{\text{x}}_{i}},{{\text{x}}_{j}})]}_{ij}}\in {{\mathbb{R}}^{m\,\times m}}
\end{align}
and 
\begin{align}
l({{\text{x}}_{i}},{{\text{x}}_{j}})=\,||{{\text{x}}_{i}}-{{\text{x}}_{j}}|{{|}_{1}},\,\forall \,i,j=1,...,m.
\end{align}

Note that $\text{D}\in {{\mathbb{R}}^{m\,\times \,m}},$ ${{\text{x}}_{i}}\,\,({{\text{x}}_{j}},$respectively) is the $i$th ($j$th, respectively) feature vector of feature matrix S, and $||{{\text{x}}_{i}}-{{\text{x}}_{j}}|{{|}_{1}}=\sum\nolimits_{a=1}^{n}{|x_{i}^{a}-x_{j}^{a}|}$ is the Manhattan distance between ${{\text{x}}_{i}}\,\,\text{and}\,\,{{\text{x}}_{j}}$. The  eigenvalues of the distance matrix D are denoted by ${{\lambda }_{1}}<{{\lambda }_{2}}<...<{{\lambda }_{m}}$, and ${{\text{v}}_{1}},{{\text{v}}_{2}},...,{{\text{v}}_{m}}$ denote the corresponding eigenvectors. According to the Courant-Fischer Theorem \cite{ouyang2016noise}, we then have
\begin{figure}[h]
\vspace*{-\baselineskip}
  \includegraphics[width=.96\linewidth]{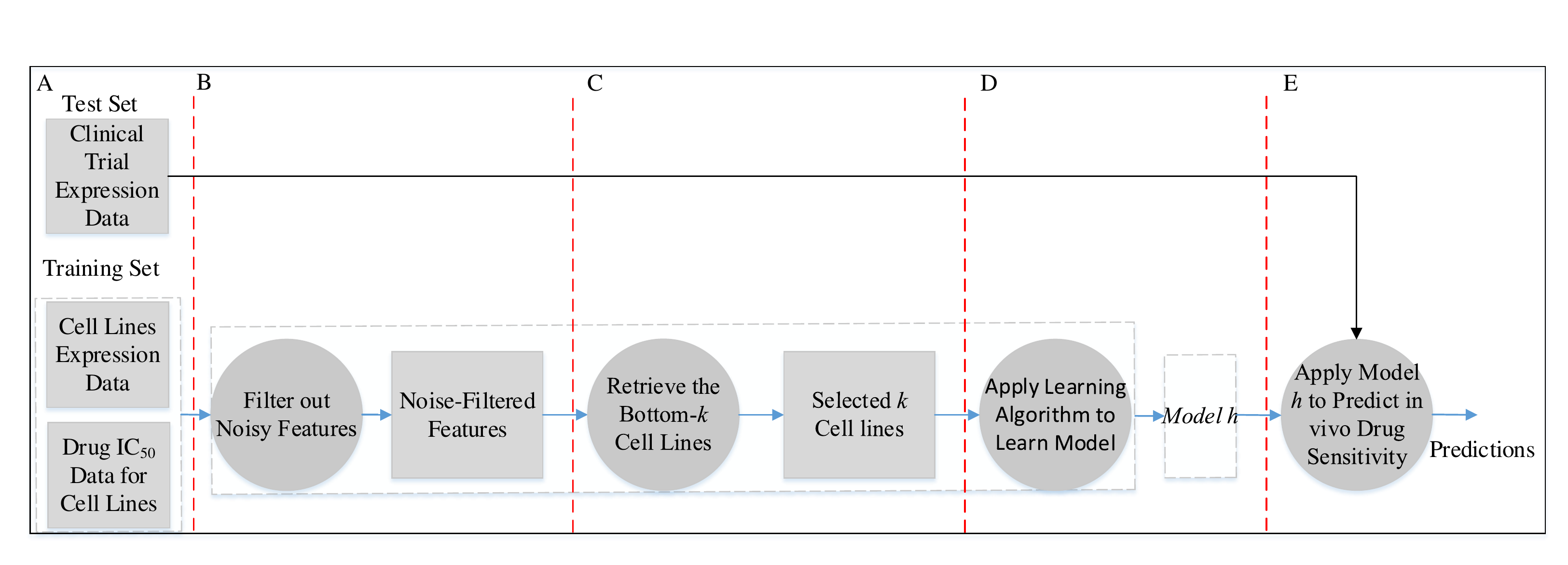}
   \vspace*{-5mm}
  \caption{Data flow diagram showing our proposed approach to predicting in vivo drug sensitivity.}
   
\end{figure}
\vspace*{-4mm}
\begin{align}
{{\text{v}}_{1}}=\,\arg \,\underset{\text{z}:||\text{z}|{{|}_{2}}=1}{\mathop{\min }}\,\,{{\text{z}}^{T}}\text{D}\,\text{z}
\end{align}
and

\begin{align}
{{\text{v}}_{l}}=\,\arg \,\underset{\text{z}:||\text{z}|{{|}_{2}}=1,\,\text{z}\,\bot \,\text{span}\{{{\text{v}}_{1}},{{\text{v}}_{2}},...,{{\text{v}}_{l-1}}\}}{\mathop{\min }}\,\,{{\text{z}}^{T}}\text{D}\,\text{z}\text{.}
\end{align}

 Let ${{\text{V}}_{t}}=[{{\text{v}}_{1}},{{\text{v}}_{2}},...,{{\text{v}}_{t}}]$ be the matrix whose columns are the first $\textit{t}$ eigenvectors of the distance matrix D with the smallest eigenvalues (in this study, $t=1.)$ We project the training input of cell lines $\text{X}$ onto ${{\text{V}}_{t}}$ to obtain the transformed training input $\overline{\text{X}}\in {{\mathbb{R}}^{m\,\times \,n}}$ with noise-filtered features. That is,
\begin{align}
\overline{\text{X}}={{\text{V}}_{t}}\text{V}_{t}^{T}\text{X}\text{.}
\end{align}
(C) Compute the degrees between each training input ${{\text{x}}_{i}}\in \text{X}$ and the corresponding transformed training input $\overline{{{\text{x}}_{i}}}\in \overline{\text{X}}$ for $\textit{i}$=1..$\textit{m}$
\begin{align}
{{i}^{*}}=\,\arg \,\underset{i\in \{1,2,...,m\}}{\mathop{\min }}\,{{\theta }_{i}}\,\text{ }
\end{align}
and
\begin{align}
{{\theta }_{i}}=\,{{\cos }^{-1}}\left( \frac{\overline{{{\text{x}}_{i}}}.\,{{\text{x}}_{i}}}{||\overline{{{\text{x}}_{i}}}|{{|}_{2}}\,||{{\text{x}}_{i}}|{{|}_{2}}} \right)\,\times \frac{180}{\pi }\text{ }\text{.}
\end{align}
In this case, ${{i}^{*}}$ is the index of ${{\text{x}}_{{{i}^{*}}}}\in \mathbf{X}$, whose degree is ${{\theta }_{{{i}^{*}}}}$. Denote by $\overline{\text{S}}=\{({{x}_{{{i}^{*}}}},{{y}_{{{i}^{*}}}})\}_{{{i}^{*}}=1}^{q}\subseteq \,\,\text{S}$ the reduced training set whose $\textit{q}$ feature vectors correspond to the training set with the smallest $q$ degrees (i.e., ${{\theta }_{{{i}^{*}}=1}}<{{\theta }_{{{i}^{*}}=2}}<...<{{\theta }_{{{i}^{*}}=q}}$ where $\textit{q < m}$). (D) A learning algorithm is called on the training set $\overline{\text{S}}$ of size $q$ to induce model $h$. (E) We apply model $h$ to make predictions on the test set. In the rest of this paper, we will refer to the prediction algorithms that employ our proposed approach using the abbreviations PA+SVR+L, PA+SVR+S, and PA+RR (see Table 1). 
\vspace{.3cm}
\begin{table}[H]
\vspace*{-\baselineskip}
\scalebox{0.75}
{
\centering
\caption{Abbreviation of drug sensitivity prediction algorithms.}
\begin{tabular}{lllllll}
\hline\noalign{\smallskip}
Abbreviation & Prediction Algorithm  \\
\noalign{\smallskip}
\hline
\noalign{\smallskip}
PA+SVR+L& The proposed approach using support vector regression with a linear kernel  \\
PA+SVR+S& The proposed approach using support vector regression with a sigmoid kernel  \\
PA+RR& The proposed approach using ridge regression \\
B+SVR+L& The baseline approach using support vector regression with a linear kernel  \\
B+SVR+S& The baseline approach using support vector regression with a sigmoid kernel  \\
B+RR&The baseline approach using ridge regression \\
\hline
\end{tabular}
}
\vspace*{-2.5mm}
\end{table}
\section{Experiments and Results}
\parskip 0pt
\subsection{Datasets}
\parskip 0pt  
The training sets correspond to an $482\text{ }\times 6539$ matrix and an $280\text{ }\times 9115$ matrix for breast cancer and multiple myeloma, respectively. The test sets correspond to an $24\text{ }\times 6538$ matrix (excluding labels) and an $188\text{ }\times 9114$ matrix (excluding labels) for breast cancer and multiple myeloma, respectively. Cell lines expression data for breast cancer and multiple myeloma were download from the ArrayExpress repository (accession number E-MTAB-783).  The drug $\text{I}{{\text{C}}_{50}}$ values for docetaxel and bortezomib were downloaded from (http://genemed.uchicago.edu/$\text{ }\!\!\tilde{\ }\!\!\text{ }$pgeeleher/cgpPrediction/). The clinical trial data for breast cancer (multiple myeloma, respectively) were downloaded from the Gene Expression Omnibus (GEO) repository with accession numbers GSE350 and GSE349 (GSE9782, respectively) \cite{edgar2002gene,chang2003gene}. The responses (i.e., labels) of the clinical trial data are categorical (e.g., “sensitive” or “resistant”). These labels were clinically evaluated by the degree of reduction in tumor size to the given drug \cite{geeleher2014clinical}. The downloaded data were processed according to Geeleher et al. \cite{geeleher2014clinical}.
\subsection{Experimental Methodology}
\parskip 0pt  
10-fold cross validation is not suitable in this study as labels of the testing sets are categorical while the labels of the corresponding training sets are real values. Hence, to evaluate whether the proposed approach exhibits stable superior performances as sample size changes, we reduced the sample size for the training set by 1\% for each run, until the reduction reached 4\%. In other words, we performed 5 runs with sample sizes of (482, 478, 473, 468, 463) and (280, 278, 275, 272, 269) for the two real datasets, respectively. 

\quad Each prediction algorithm was trained on the same given training set, whose labels were continuous, to yield models. Then, each model was applied to the same test set to yield predictions, where the accuracy of the prediction algorithms were measured using the $\textit{Area Under the ROC Curve}$ (AUC) as in \cite{geeleher2014clinical}.
\normalsize

\quad The software used in this work included support vector regression with linear and sigmoid kernels in the LIBSVM package \cite{chang2011libsvm}, ridge regression \cite{geeleher2014clinical}, and R code for processing the datasets and performance evaluation \cite{geeleher2014clinical}. We used R to write code for the prediction algorithms and to perform the experiments. 
\subsection{Experimental Results}
\normalsize
Table 2 and Table 3 show the AUCs of 6 docetaxel and bortezomib sensitivity prediction algorithms on the clinical trial data for breast cancer and multiple myeloma, respectively. Columns “$\textit{m}$” and “$\textit{d}$” show the number of cell lines and genes, respectively, in the training sets that were provided to each prediction algorithm; we provided the same training sets to each prediction algorithm. Column “$\textit{q}$+PA” shows the number of selected (i.e., retrieved) cell lines that were used in PA+SVR+L, PA+SVR+S, and PA+RR to learn the models.

\quad Table 2 and Table 3 show that our prediction algorithms perform better than the baseline prediction algorithms (i.e., B+SVR+L and B+SVR+S) and B+RR, the proposed prediction algorithm by Geeleher et al \cite{geeleher2014clinical}. The results are dominant compared to the other prediction algorithms on the clinical trial data in terms of the AUC of each run and of MAUC. These results indicate the stability of the proposed prediction algorithms.

\begin{table}[H]
\scalebox{0.6}
{
\centering
\caption{The AUC of docetaxel sensitivity prediction algorithms for breast cancer patients on the test set. The algorithm with the highest AUC is shown in bold. MAUC is the mean AUC.}
\begin{tabular}{llllllllllll}
\hline\noalign{\smallskip}
\textit{m}&$\textit{d}$&PA+SVR+L&PA+SVR+S&PA+RR&$\textit{q}$+PA&B+SVR+L&B+SVR+S&B+RR\\
\noalign{\smallskip}
\hline
\noalign{\smallskip}
482&6,538&0.842&\textbf{0.878}&0.821&478&0.835&0.842&0.814\\
478&6,538&0.807&\textbf{0.871}&0.814&474&0.814&\textbf{0.871}&0.814\\
473&6,538&0.814&\textbf{0.878}&0.821&469&0.800&0.864&0.821\\
468&6,538&0.828&\textbf{0.864}&0.828&464&0.821&0.857&0.821\\
463&6,538&0.857&\textbf{0.864}&0.821&459&0.835&0.857&0.821\\
\hline
\noalign{\smallskip}
MAUC&-&0.829&0.871&0.821&-&0.821&0.858&0.818\\

\hline
\end{tabular}
}
\vspace*{-2mm}
\end{table}

\begin{table}[H]
\vspace*{-\baselineskip}
\scalebox{0.6}
{
\centering

\caption{The AUC of bortezomib sensitivity prediction algorithms for multiple myeloma patients on the test set. The algorithm with the highest AUC is shown in bold. The MAUC is the mean AUC.}
\begin{tabular}{llllllllllll}
\hline\noalign{\smallskip}
$\textit{m}$&$\textit{d}$&PA+SVR+L&PA+SVR+S&PA+RR&$\textit{q}$+PA&B+SVR+L&B+SVR+S&B+RR\\
\noalign{\smallskip}
\hline
\noalign{\smallskip}
280&9,114&\textbf{0.659}&0.635&0.654&210&0.613&0.602&0.614\\
278&9,114&\textbf{0.656}&0.623&0.650&209&0.609&0.600&0.611\\
275&9,114&\textbf{0.679}&0.626&0.647&207&0.622&0.601&0.603\\
272&9,114&\textbf{0.685}&0.641&0.653&204&0.628&0.605&0.607\\
269&9,114&\textbf{0.681}&0.658&0.657&202&0.632&0.598&0.606\\
\hline
\noalign{\smallskip}
MAUC&-&0.672&0.636&0.652&-&0.620&0.601&0.608\\

\hline
\end{tabular}
}
\vspace*{-2mm}
\end{table}

Figure 2 and Figure 3 show the predictions of three prediction algorithms on the test sets when the prediction algorithms learned models from the training sets with sizes $\textit{m}$=482 and $\textit{m}$=280, respectively. For PA+SVR+S in Figure 2(a), the difference between the predicted drug sensitivity in breast cancer patients was highly statistically significant ($P=67\,\times \,{{10}^{-5}}$ from a $\textit{t}$-test) between trial-defined sensitivity and resistant groups.  The B+SVR+S (Figure 2(b)) and B+RR (Figure 2(c)) achieved significant results with $P= 117\,\times \,{{10}^{-5}}$ and $P=434\,\times \,{{10}^{-5}}$  , respectively, from $\textit{t}$-tests.  PA+SVR+L (Figure 3(a)) achieved highly significant results ($P=10\,\times \,{{10}^{-5}}$ from a $\textit{t}$-test). The results of B+RR (Figure 3(b)) and B+SVR+L (Figure 3(c)) were also significant with $P=261\,\times \,{{10}^{-5}}$ and $P=556\,\times \,{{10}^{-5}}$, respectively, from $\textit{t}$-tests.

\begin{figure}[h]

 \begin{framed}

  \begin{minipage}{\linewidth}
  \raggedright
  \hspace{-.28cm}
  \includegraphics[width=.26\linewidth]{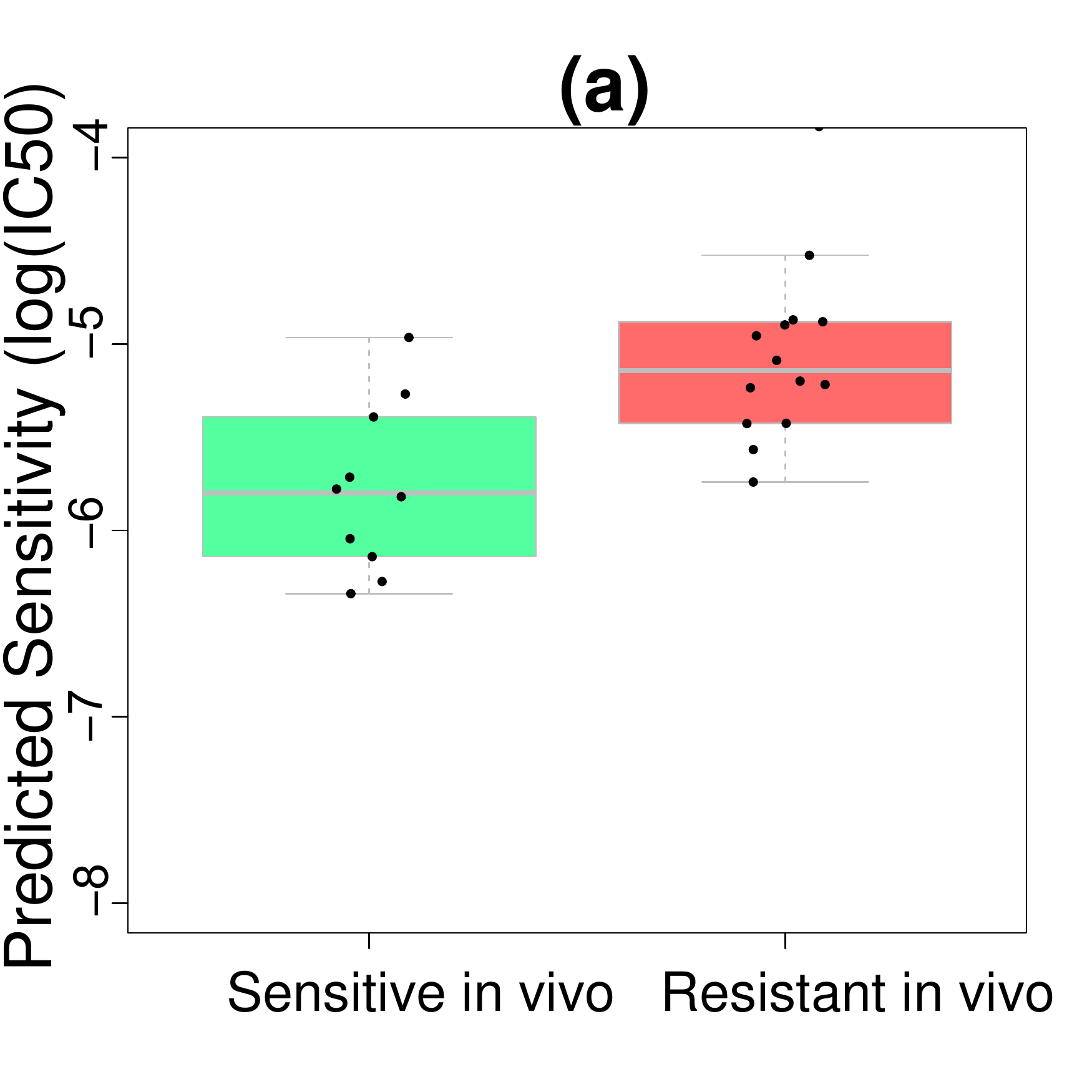}\hspace{-.28cm}
  \includegraphics[width=.26\linewidth]{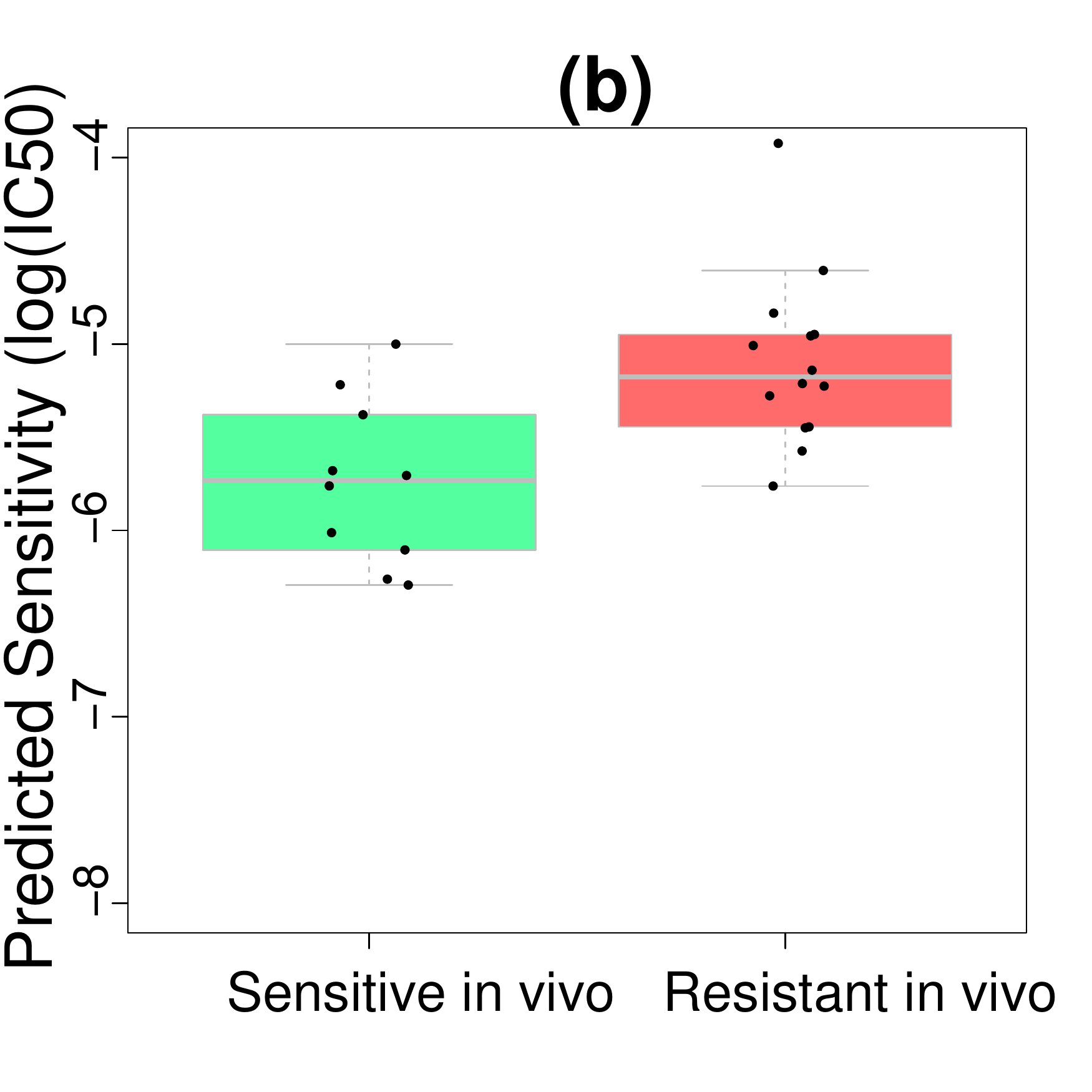}\hspace{-.27cm}
   \includegraphics[width=.26\linewidth]{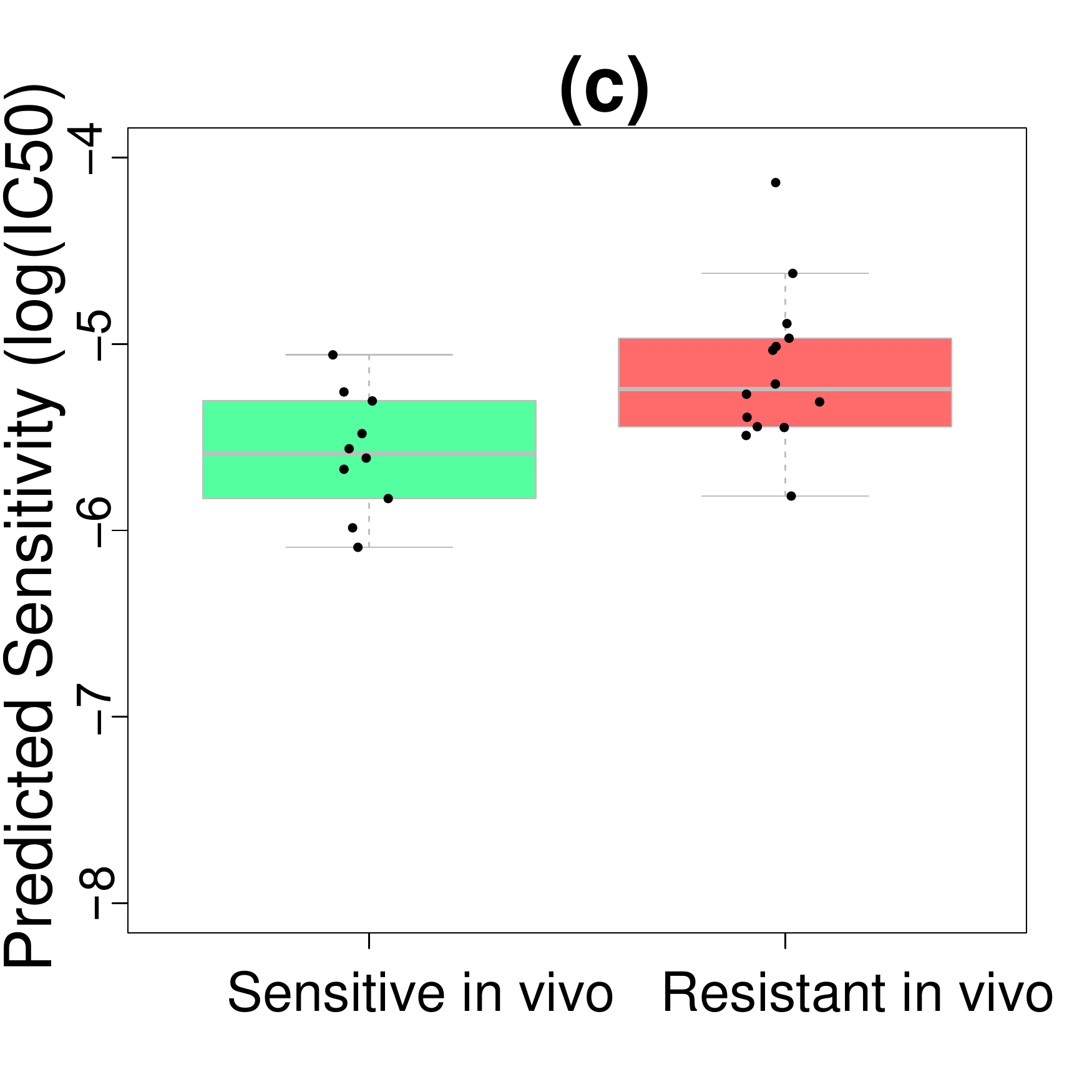}\hspace{-.27cm}
  \includegraphics[width=.26\linewidth]{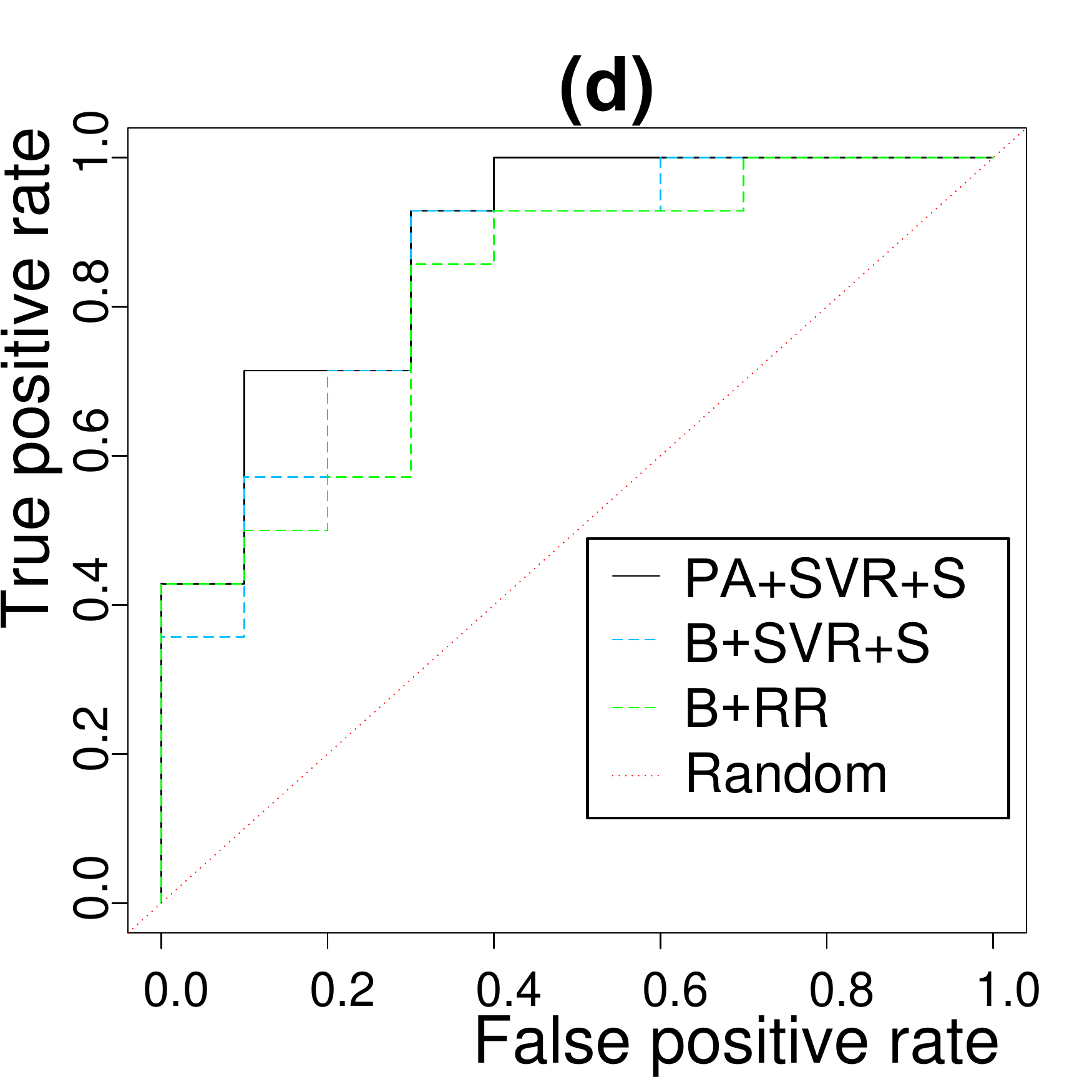}\hspace{-.27cm}
   \parbox[c]{\textwidth}{\caption*{} Figure 2: Predictions of docetaxel sensitivity in breast cancer patients.  Strip charts (a), (b), and (c) show the differences in predicted drug sensitivity for individuals sensitive or resistant to docetaxel treatment using PA+SVR+S, B+SVR+S, and B+RR prediction algorithms, while (d) shows the ROC curves of the prediction algorithms, which reveal the proportion of true positives compared to the proportion of false positives. ROC = receiver operating characteristic.}
  \end{minipage}%
  \end{framed}
\end{figure}
\vspace{-1cm} 
\begin{figure}[h]
 \begin{framed}

  \begin{minipage}{\linewidth}
  \includegraphics[width=.26\linewidth]{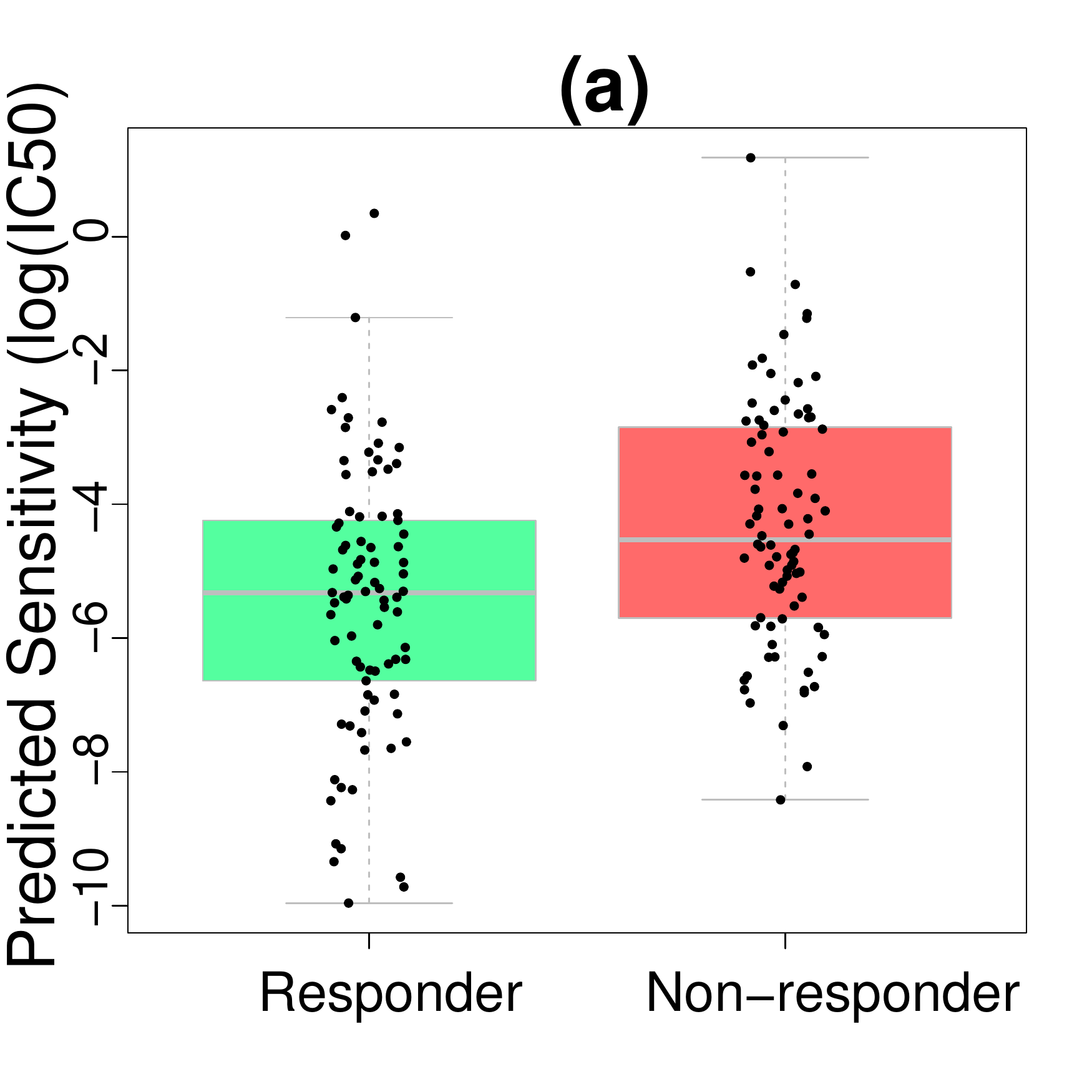}\hspace{-.25cm}
  \includegraphics[width=.26\linewidth]{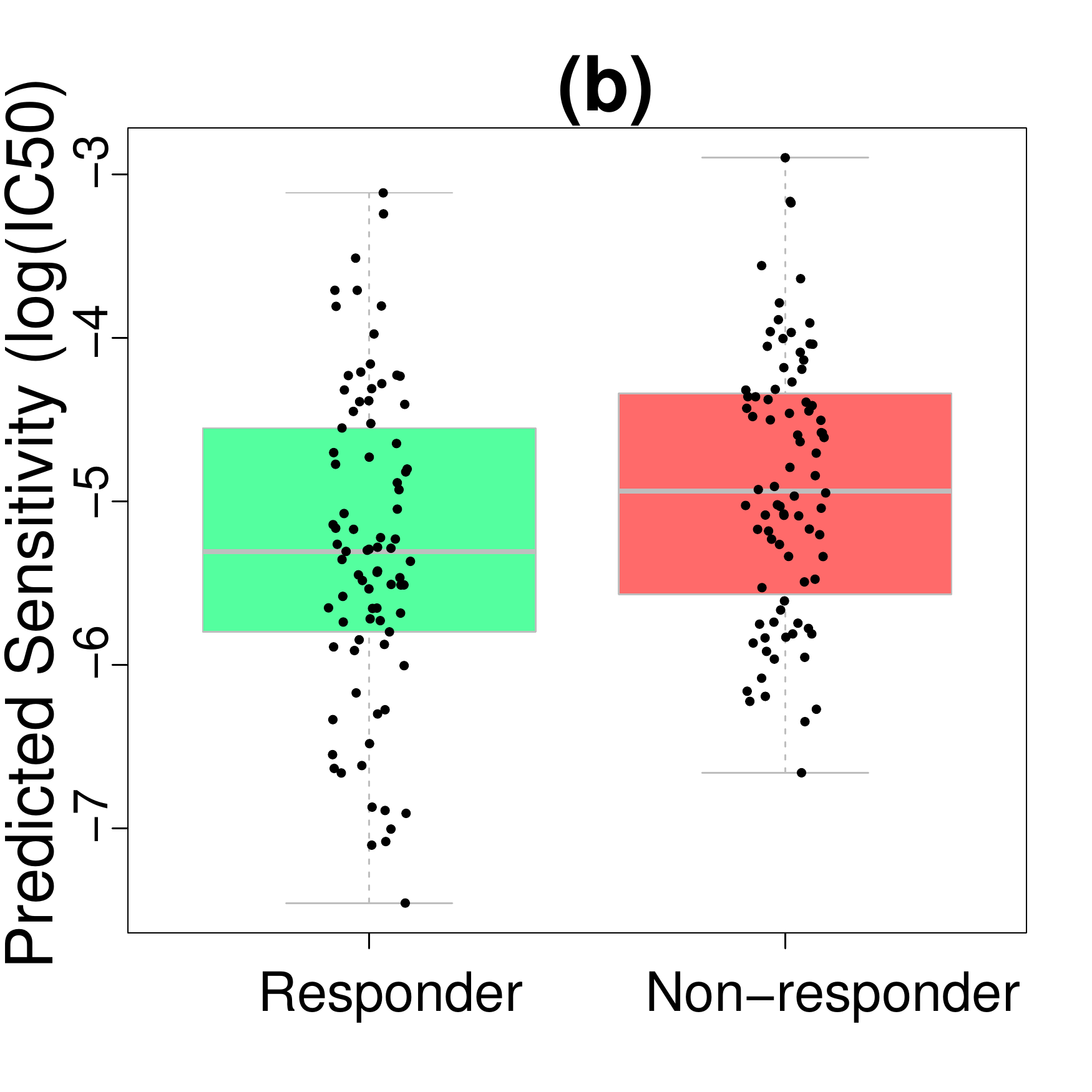}\hspace{-.25cm}
  \includegraphics[width=.26\linewidth]{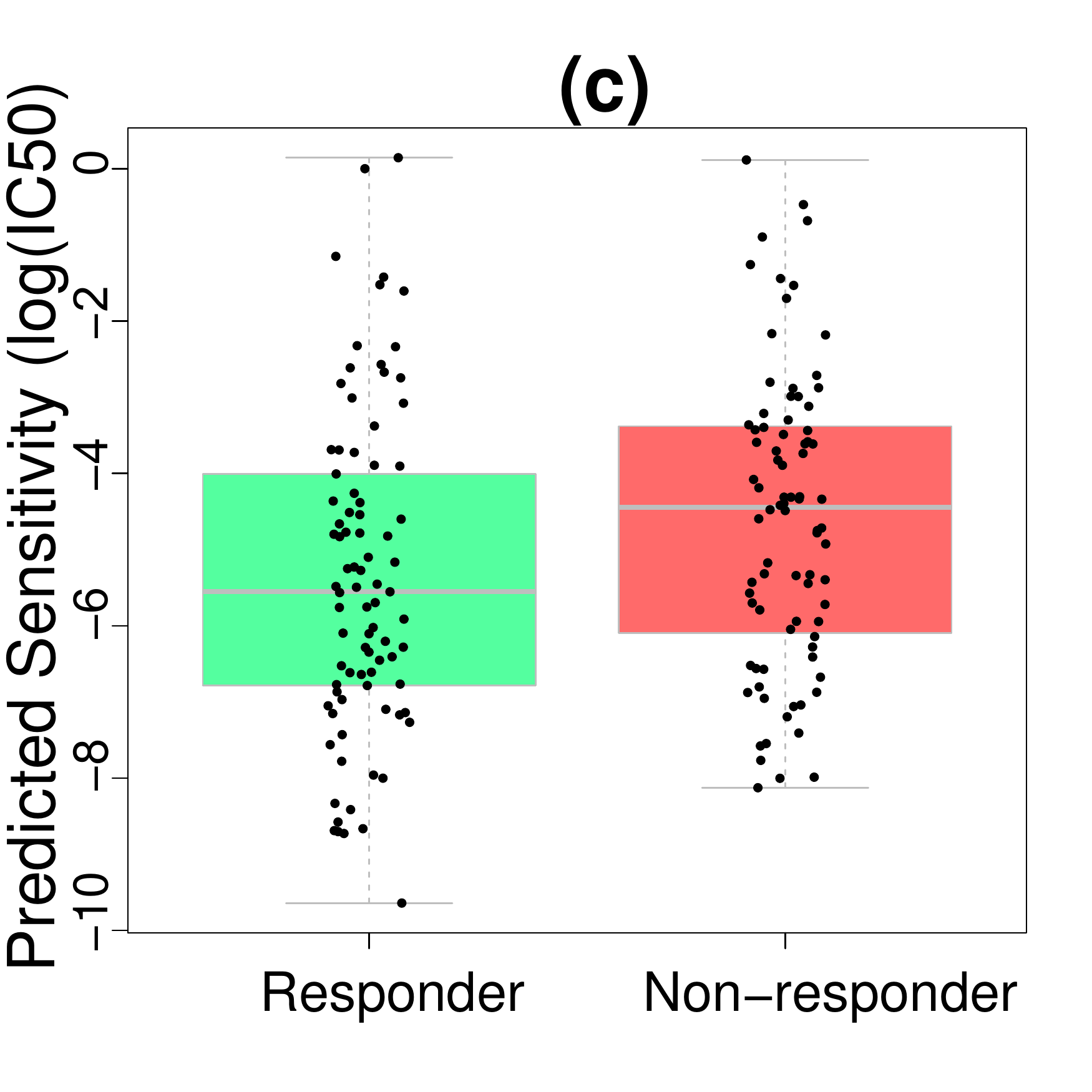}\hspace{-.25cm}
  \includegraphics[width=.26\linewidth]{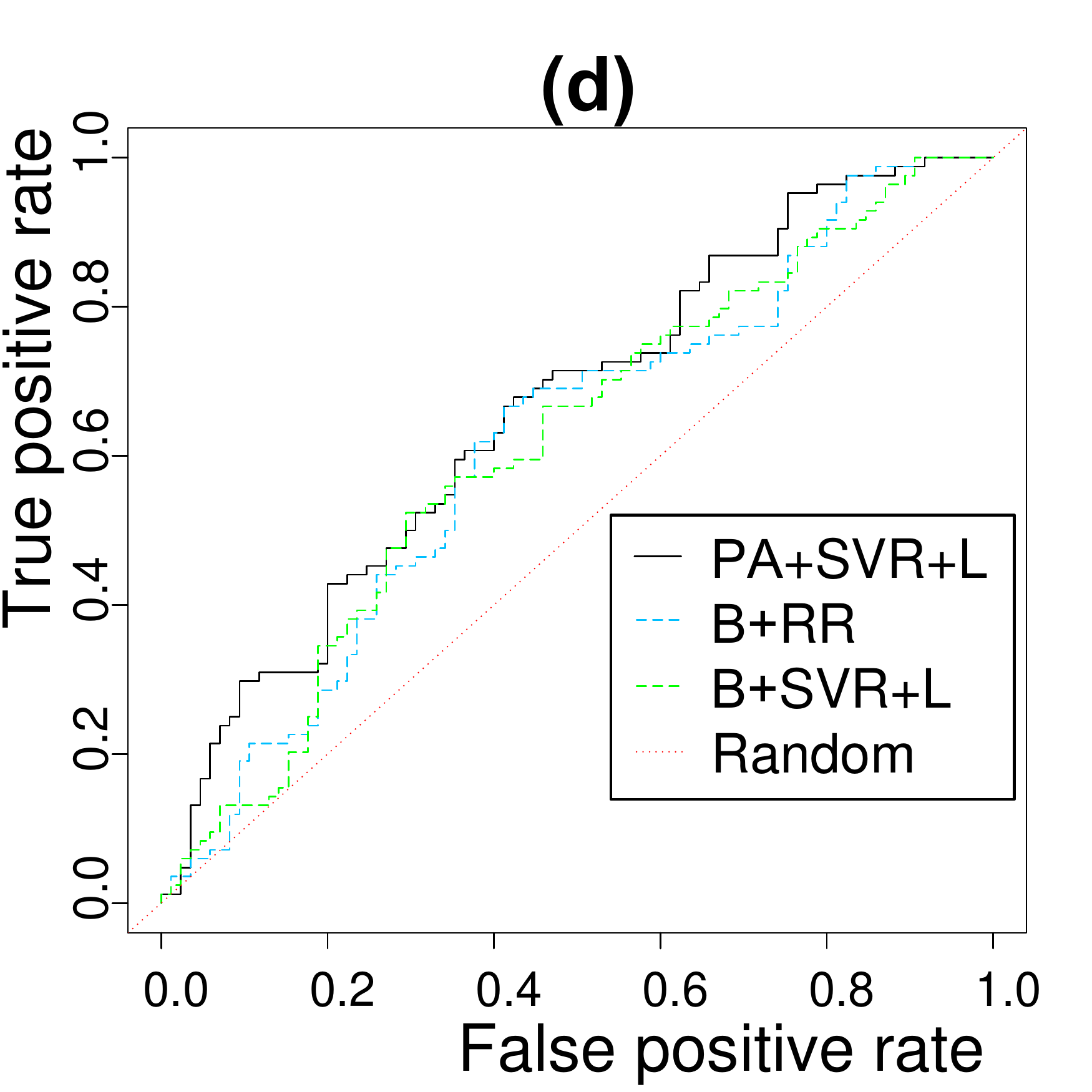}\hspace{-.25cm}
  \parbox[c]{\textwidth}{\caption*{} Figure 3: Prediction of bortezomib sensitivity in multiple myeloma patients. The strip charts and boxplots in (a), (b), and (c) show predicted drug sensitivity for in vivo responders and non-responders to bortezomib using the PA+SVR+L, B+RR, and B+SVR+L prediction algorithms, respectively.  The ROC curve in (d) illustrates the estimated prediction accuracy of the prediction algorithms.}
  \end{minipage}%



\end{framed}
\end{figure}
\vspace{-1cm}
\section{Conclusion}
\parskip 0pt  
\normalsize
In this paper, we proposed a noise-filtering approach that blends the following techniques: (1) Numerical linear algebra to yield transformed training input corresponding to noise-filtered features; (2) Information retrieval to retrieve better quality cancer cell lines from the training set according to the minimum degrees between the training input and the transformed training input; (3) Machine learning to learn a model from better quality, reduced-size training sets and perform predictions on test sets. The proposed approach uses two machine learning algorithms: support vector regression and ridge regression. Compared to an existing drug sensitivity approach, our proposed approach results in higher and statistically significant AUC values when using clinical trial data.

\bibliography{nips_2016}

\begin{thebibliography}{10}

\bibitem{siegel2015cancer}
Rebecca~L Siegel, Kimberly~D Miller, and Ahmedin Jemal.
\newblock Cancer statistics, 2015.
\newblock {\em CA: a cancer journal for clinicians}, 65(1):5--29, 2015.

\bibitem{amcstarver2016cancer}
Cancer facts \& figures 2016. {A}merican {C}ancer {S}ociety.

\bibitem{kamb2007cancer}
Alexander Kamb, Susan Wee, and Christoph Lengauer.
\newblock Why is cancer drug discovery so difficult?
\newblock {\em Nature Reviews Drug Discovery}, 6(2):115--120, 2007.

\bibitem{marx2015cancer}
Vivien Marx.
\newblock Cancer: a most exceptional response.
\newblock {\em Nature}, 520(7547):389--393, 2015.

\bibitem{bedard2013tumour}
Philippe~L Bedard, Aaron~R Hansen, Mark~J Ratain, and Lillian~L Siu.
\newblock Tumour heterogeneity in the clinic.
\newblock {\em Nature}, 501(7467):355--364, 2013.

\bibitem{roden2002genetic}
Dan~M Roden and Alfred~L George~Jr.
\newblock The genetic basis of variability in drug responses.
\newblock {\em Nature Reviews Drug Discovery}, 1(1):37--44, 2002.

\bibitem{libbrecht2015machine}
Maxwell~W Libbrecht and William~Stafford Noble.
\newblock Machine learning applications in genetics and genomics.
\newblock {\em Nature Reviews Genetics}, 16(6):321--332, 2015.

\bibitem{covell2015data}
David~G Covell.
\newblock Data mining approaches for genomic biomarker development:
  applications using drug screening data from the cancer genome project and the
  cancer cell line encyclopedia.
\newblock {\em PloS one}, 10(7):e0127433, 2015.

\bibitem{7591437embc2016}
T.~Turki and Z.~Wei.
\newblock Learning approaches to improve prediction of drug sensitivity in
  breast cancer patients.
\newblock In {\em 2016 38th Annual International Conference of the IEEE
  Engineering in Medicine and Biology Society (EMBC)}, pages 3314--3320, Aug
  2016.

\bibitem{costello2014community}
James~C Costello, Laura~M Heiser, Elisabeth Georgii, Mehmet G{\"o}nen,
  Michael~P Menden, Nicholas~J Wang, Mukesh Bansal, Petteri Hintsanen,
  Suleiman~A Khan, John-Patrick Mpindi, et~al.
\newblock A community effort to assess and improve drug sensitivity prediction
  algorithms.
\newblock {\em Nature biotechnology}, 32(12):1202--1212, 2014.

\bibitem{geeleher2014clinical}
Paul Geeleher, Nancy~J Cox, and R~Stephanie Huang.
\newblock Clinical drug response can be predicted using baseline gene
  expression levels and in vitro drug sensitivity in cell lines.
\newblock {\em Genome biology}, 15(3):1, 2014.

\bibitem{yadav2015drug}
Bhagwan Yadav, Peddinti Gopalacharyulu, Tea Pemovska, Suleiman~A Khan,
  Agnieszka Szwajda, Jing Tang, Krister Wennerberg, and Tero Aittokallio.
\newblock From drug response profiling to target addiction scoring in cancer
  cell models.
\newblock {\em Disease Models and Mechanisms}, 8(10):1255--1264, 2015.

\bibitem{mohri2012foundations}
Mehryar Mohri, Afshin Rostamizadeh, and Ameet Talwalkar.
\newblock {\em Foundations of machine learning}.
\newblock MIT press, 2012.

\bibitem{Manning:2008:IIR:1394399}
Christopher~D. Manning, Prabhakar Raghavan, and Hinrich Sch\"{u}tze.
\newblock {\em Introduction to Information Retrieval}.
\newblock Cambridge University Press, New York, NY, USA, 2008.

\bibitem{scholkopf2002learning}
Bernhard Sch{\"o}lkopf and Alexander~J Smola.
\newblock {\em Learning with kernels: support vector machines, regularization,
  optimization, and beyond}.
\newblock MIT press, 2002.

\bibitem{breiman2001random}
Leo Breiman.
\newblock Random forests.
\newblock {\em Machine learning}, 45(1):5--32, 2001.

\bibitem{ouyang2016noise}
Bo~Ouyang, Lurong Jiang, and Zhaosheng Teng.
\newblock A noise-filtering method for link prediction in complex networks.
\newblock {\em PloS one}, 11(1):e0146925, 2016.

\bibitem{edgar2002gene}
Ron Edgar, Michael Domrachev, and Alex~E Lash.
\newblock Gene expression omnibus: Ncbi gene expression and hybridization array
  data repository.
\newblock {\em Nucleic acids research}, 30(1):207--210, 2002.

\bibitem{chang2003gene}
Jenny~C Chang, Eric~C Wooten, Anna Tsimelzon, Susan~G Hilsenbeck, M~Carolina
  Gutierrez, Richard Elledge, Syed Mohsin, C~Kent Osborne, Gary~C Chamness,
  D~Craig Allred, et~al.
\newblock Gene expression profiling for the prediction of therapeutic response
  to docetaxel in patients with breast cancer.
\newblock {\em The Lancet}, 362(9381):362--369, 2003.

\bibitem{chang2011libsvm}
Chih-Chung Chang and Chih-Jen Lin.
\newblock Libsvm: a library for support vector machines.
\newblock {\em ACM Transactions on Intelligent Systems and Technology (TIST)},
  2(3):27, 2011.

\end{thebibliography}
\bibliographystyle{unsrt}
\end{document}